\documentclass{article} 
\usepackage{iclr2027_conference,times}

\usepackage{amsmath,amsfonts,bm}

\def\eqref#1{equation~\ref{#1}}

\def\1{\bm{1}}

\DeclareMathAlphabet{\mathsfit}{\encodingdefault}{\sfdefault}{m}{sl}
\SetMathAlphabet{\mathsfit}{bold}{\encodingdefault}{\sfdefault}{bx}{n}

\usepackage{amsthm,amssymb,xcolor}
\usepackage{graphicx}
\usepackage{booktabs}
\usepackage{tabularx}
\usepackage{colortbl}
\usepackage{float}
\usepackage{subcaption}
\usepackage{hyperref}
\usepackage{url}

\theoremstyle{plain}
\newtheorem{lemma}{Lemma}[section]
\newtheorem{proposition}[lemma]{Proposition}
\definecolor{InsightBlue}{RGB}{35,88,155}
\newcommand{\insightbox}[2]{%
  \par\smallskip\noindent
  \begingroup
  \setlength{\fboxsep}{7pt}%
  \setlength{\fboxrule}{0.7pt}%
  \fcolorbox{InsightBlue}{InsightBlue!4}{%
    \parbox{\dimexpr\linewidth-2\fboxsep-2\fboxrule\relax}{%
      \textcolor{InsightBlue}{\textbf{Insight #1.}}\ #2%
    }%
  }%
  \endgroup
  \par\smallskip
}

\title{ESEA: Error-Aware Sparse Expert Allocation for Efficient Diffusion MoE Inference}

\author{%
Jing Xu\textsuperscript{1,*},
Jiateng Wei\textsuperscript{2,*},
Chenqian Yan\textsuperscript{1,*},
Songwei Liu\textsuperscript{2,\S}\\[0.4em]
\textsuperscript{1}Xiamen University \qquad
\textsuperscript{2}Zhejiang University\\[0.25em]
{\small \textsuperscript{*}Equal contribution \qquad
\textsuperscript{\S}Corresponding author}
}

\iclrfinalcopy 
\begin{document}

\maketitle

\begin{abstract}
Mixture-of-Experts (MoE) has emerged as a promising paradigm for scaling diffusion models, yet activating a fixed number of experts for all tokens leaves computational redundancy. Existing expert skipping methods mainly target LLMs and overlook the iterative denoising nature of diffusion models, leading to degraded generation quality when directly applied to diffusion moe models. We analyze the underlying error mechanisms and identify three insights: (i) approximation error depends on expert similarity; (ii) local errors exhibit timestep–layer heterogeneity and cross-prompt consistency; (iii) the impacts of local errors on global error vary across timesteps. Based on these insights, we propose ESEA (\textbf{E}rror-aware \textbf{S}parse \textbf{E}xpert \textbf{A}llocation), a training-free method for diffusion MoE inference. ESEA first adjusts routing weights based on expert similarity to reduce approximation error, then weights calibrated 3D local errors tensor to reflect their timestep-dependent impacts on global error. Finally, dynamic programming minimizes the weighted local error, jointly allocating experts across timesteps and layers under a fixed total expert budget. The routing weight adjustment strategy and expert allocation are determined offline, introducing negligible overhead during inference. Extensive experiments on HunyuanImage-3.0-Instruct, its distilled variant, and LingBot-Video demonstrate that ESEA achieves the best performance among the compared methods under the same expert budget.
\end{abstract}

\begin{figure}[H]
\centering
\includegraphics[width=\linewidth]{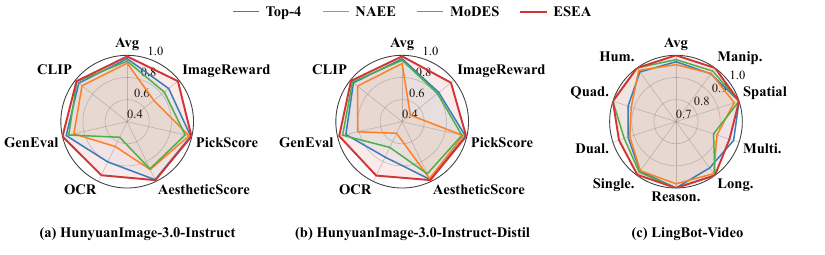}
\caption{Comparison of expert skipping methods across three diffusion MoE models at an average Top-$k$ budget of 4. Scores are normalized to the corresponding no expert skipping baseline.}
\label{fig:main-results-radar}
\end{figure}

\section{Introduction}


Diffusion models have made substantial progress in image~\citep{cai2025z,zhao2026qwen,mao2026wan} and video generation~\citep{seedance2026seedance,zhu2026dreamx,chen2026skyreels}, enabling high-quality visual content synthesis while highlighting the need to scale model capacity efficiently as generation tasks become increasingly complex.
Mixture-of-Experts (MoE) architectures have been widely adopted in large language models (LLMs)~\citep{yang2025qwen3,xu2026deepseek,team2026kimi,xiao2026mimo,zeng2026glm}, offering an effective approach to balancing model capacity and computational cost through sparse conditional computation.
Specifically, MoE organizes feed-forward computation into multiple independent experts and uses a lightweight router to activate only a small subset for each token, allowing the total parameter count to grow without a proportional increase in per-token computation.
Recent work has extended this paradigm to diffusion-based generative models~\citep{team2025hunyuanimage,ma2026scaling,shi2026mamoda2}, demonstrating its potential for scaling visual generation efficiently.
For example, LingBot-Video~\citep{ma2026scaling}, with 30B total parameters and 3B active parameters, approaches the training loss of a 14B dense model while achieving a $2.59\times$ inference speedup over a depth-matched 14B dense baseline at a sequence length of 1M tokens.


Despite these advantages, standard MoE routing activates a fixed number of experts for every token, incurring unnecessary computation and limiting inference efficiency.
To mitigate this issue, prior work~\citep{lu2024not,huang2024harder,huang2025mixture,zeng2024adamoe,huang2026modes,wu2026sere} has proposed Dynamic Expert Skipping methods that adaptively reduce the number of active experts during inference.
However, directly transferring these methods, which are primarily designed for LLMs and vision-language models (VLMs), to diffusion models can severely degrade generation quality, as shown in Figure~\ref{fig:main-results-radar}, largely because they do not adequately account for the iterative nature of denoising and the computational demands that vary across timesteps.


Diffusion models generate visual content through iterative denoising, typically progressing from global structure formation to local detail refinement as the noise level decreases.
Feature distributions, routing behaviors, and expert computation requirements evolve throughout this process, meaning that the appropriate number of active experts may vary substantially across timesteps, even within the same network layer.
Consequently, ignoring this variation can lead to overly aggressive expert skipping at sensitive stages, compromising generation quality, while retaining excessive computation at more redundant stages and limiting acceleration.
Local errors introduced by expert skipping can propagate through subsequent denoising steps, potentially accumulating and becoming amplified along the trajectory.
In addition, diffusion inference follows a prefill-only computation pattern, making methods~\citep{oncescu2025opportunistic,vankov2026xshare,wu2026sere} designed to accelerate the decoding stage in LLMs and VLMs inapplicable to this setting.


To address this problem, we first conduct an in-depth analysis of the error mechanisms underlying dynamic expert skipping in diffusion MoE models and identify three key insights corresponding to approximation error, local error, and global error:
(i) The approximation error introduced by Top-$k$ reduction arises from using retained experts to approximate discarded experts and depends on expert similarity. 
(ii) Local errors exhibit pronounced timestep--layer heterogeneity and maintain consistent patterns across prompts. Specifically, the same Top-$k$ configuration can induce substantially different local errors across denoising timesteps and layers.
(iii) Local errors at different timesteps have different impacts on global error, reflecting their timestep-dependent importance. Even local errors of the same magnitude can affect the final output differently, depending on when they are introduced and how they propagate through subsequent denoising steps.

To this end, we propose ESEA (\textbf{E}rror-aware \textbf{S}parse \textbf{E}xpert \textbf{A}llocation),a training-free and error-aware expert allocation method for diffusion MoE inference.
In response to the first insight, we use the precomputed similarity to determine whether the retained experts' routing weights are rescaled to compensate for discarded contributions.
To capture timestep--layer heterogeneity, we construct a three-dimensional local error tensor over denoising timesteps, network layers, and candidate Top-$k$ configurations through offline calibration, and formulate it as a discrete allocation problem under a fixed total expert budget.
To account for the timestep-dependent impact of local errors on global error, ESEA further weights these errors according to the noise level at each timestep,aligning with the denoising process.
By solving this problem with dynamic programming, we obtain the Top-$k$ configuration that minimizes the errors under the given budget, reducing expert computation while controlling generation quality loss.
At inference time, the Top-$k$ configuration and the the routing weight
adjustment strategy for each timestep and layer are retrieved from the precomputed configuration table, introducing negligible computational overhead.

In summary, our key contributions are as follows:
\begin{itemize}
    \item We systematically analyze the error mechanisms underlying dynamic expert skipping in diffusion MoE models, revealing the role of expert similarity in approximation error, pronounced timestep--layer heterogeneity and cross-prompt consistency in local errors, and the timestep-dependent impacts of local errors on global error.
    \item We propose ESEA (\textbf{E}rror-aware \textbf{S}parse \textbf{E}xpert \textbf{A}llocation), a training-free and error-aware expert allocation method for diffusion MoE inference.
    ESEA reduces approximation error through similarity-based routing weight adjustment and minimizes timestep-weighted local errors via dynamic programming, jointly allocating experts across timesteps and layers under a fixed total expert budget.
    \item Comprehensive empirical evaluation across HunyuanImage3.0-Instruct, its distilled variant, and LingBot-Video demonstrates that ESEA achieves the best generation quality among the compared methods under the same expert budget.
\end{itemize}

\section{Related Work}

%

\textbf{Efficient MoE.} Existing work on efficient MoE models can be broadly divided into static model compression and dynamic expert skipping.
Static compression reduces model size by pruning redundant experts~\citep{lu2024not} or merging experts with similar functions~\citep{chen2024retraining}.
Quantization and low-rank decomposition~\citep{hu2025moequant,li2025structured} further reduce the memory required to store expert weights.
While effective in reducing model size, these methods often depend on calibration data or task-specific priors.
The removed experts and lost parameter information cannot be restored during inference, which may limit model capacity and generalization.

Dynamic expert skipping improves inference efficiency by reducing unnecessary expert computation.
NAEE~\citep{lu2024not} and Top-$p$ routing~\citep{huang2024harder} select experts using routing weights and cumulative probabilities, respectively.
MC-MoE~\citep{huang2025mixture} preserves expert computation for important tokens, while AdaMoE~\citep{zeng2024adamoe} introduces null experts to vary the number of active experts.
MoDES~\citep{huang2026modes} combines routing weights and layer sensitivity with modality-specific skipping thresholds, whereas SERE~\citep{wu2026sere} reroutes tokens to similar experts to reduce distinct expert accesses.
These methods primarily target LLMs and VLMs, overlooking error propagation during iterative denoising.
In contrast, ESEA reduces approximation error and jointly allocates experts across timesteps and layers through local and global error modeling under a fixed total expert budget.

\section{Preliminary}


\textbf{Mixture-of-Experts (MoE).} MoE layers implement conditional computation through multiple parallel experts, enabling sparse computation by selecting only a small subset of experts for each token.
Let the $l$-th MoE layer contain $M$ routed experts, denoted by $\mathcal{E}^{(l)}=\{E_1^{(l)},E_2^{(l)},\ldots,E_M^{(l)}\}$, each implemented as an independent multilayer perceptron (MLP).
Given an input token representation $x^{(l)}\in\mathbb{R}^{d}$, where $d$ denotes the hidden dimension, a lightweight router $\mathcal{R}^{(l)}$ produces routing logits $r^{(l)}=\mathcal{R}^{(l)}(x^{(l)})\in\mathbb{R}^{M}$ and applies softmax to obtain expert routing probabilities:
\begin{equation}
\pi_m^{(l)}
=
\frac{\exp\!\left(r_m^{(l)}\right)}
{\sum_{j=1}^{M}\exp\!\left(r_j^{(l)}\right)},
\qquad m=1,\ldots,M,
\label{eq:moe-routing-probabilities}
\end{equation}
where $\pi_m^{(l)}$ denotes the routing probability assigned to expert $E_m^{(l)}$ for the current token.
By default, the model selects $K$ experts for each token, with $\mathcal{S}_{K}^{(l)}$ denoting the index set of the $K$ experts with the highest routing probabilities.
Top-$K$ routing normalizes the probabilities of the selected experts, yielding the following aggregation weights:
\begin{equation}
w_m^{(l)}
=
\frac{\pi_m^{(l)}}
{\sum_{j\in\mathcal{S}_{K}^{(l)}}\pi_j^{(l)}},
\qquad m\in\mathcal{S}_{K}^{(l)}.
\label{eq:moe-aggregation-weights}
\end{equation}
The outputs of the selected experts are aggregated using these weights to obtain the routed expert output under the default Top-$K$ configuration:
\begin{equation}
y^{(l+1)}
=
\sum_{m\in\mathcal{S}_{K}^{(l)}}
w_m^{(l)}E_m^{(l)}\!\left(x^{(l)}\right).
\label{eq:moe-output}
\end{equation}
By activating only a small subset of experts for each token, MoE can scale its total parameter count while keeping per-token expert computation low.

\section{Error Analysis of Dynamic Expert Skipping}
\label{sec:error-analysis}


To motivate our error-aware expert allocation strategy, we analyze dynamic expert skipping in diffusion MoE models from three perspectives: approximation error, local error, and global error.
We establish a theoretical link between expert similarity and approximation error, characterize the timestep--layer heterogeneity of local error, and examine how these local errors propagate through iterative denoising and contribute to global error in the final generated output.

\subsection{Expert Similarity and Approximation Error}
\label{sec:expert-approximation-error}


Consider reducing the number of active experts from $K$ to $k$, where $1\leq k<K$.
We omit layer indices for simplicity and denote each expert output by $e_m=E_m(x)$.
Let $\mathcal{S}_k\subset\mathcal{S}_K$ and $\mathcal{D}_k=\mathcal{S}_K\setminus\mathcal{S}_k$ denote the retained and discarded expert indices, respectively.
The original output in Eq.~(\ref{eq:moe-output}) can then be decomposed as
\begin{equation}
y
=\sum_{i\in\mathcal{S}_k}w_i e_i
+\sum_{j\in\mathcal{D}_k}w_j e_j.
\label{eq:moe-output-decomposition}
\end{equation}
Under Top-$k$ routing, the discarded contributions are removed and the retained weights are normalized as in Eq.~(\ref{eq:moe-aggregation-weights}).
This normalization can be viewed as transferring the discarded routing weight to the retained experts and using their outputs to approximate those of the discarded experts, as formalized below.

\begin{samepage}
\begin{proposition}[Expert Weight Redistribution]
\label{prop:expert-weight-redistribution}
Let $\rho_k=\sum_{i\in\mathcal{S}_k}w_i$ and $\delta_k=\sum_{j\in\mathcal{D}_k}w_j$ be the total routing weights of the retained and discarded experts, respectively.
With $\gamma_{ij}=w_iw_j/\rho_k$ denoting the routing weight transferred from discarded expert $j$ to retained expert $i$, the output $y'$ under reduced Top-$k$ routing can be expressed as
\begin{equation}
y'
=\sum_{i\in\mathcal{S}_k}
\left(w_i e_i+\sum_{j\in\mathcal{D}_k}\gamma_{ij}e_i\right).
\label{eq:expert-redistribution-output}
\end{equation}
\end{proposition}
\end{samepage}

The detailed proof of Proposition~\ref{prop:expert-weight-redistribution} is provided in Appendix~\ref{app:expert-weight-redistribution-proof}.
This proposition shows that Top-$k$ reduction redistributes the discarded routing weight among the retained experts in proportion to their original weights.
For a retained expert $i$, each discarded expert $j$ transfers a fraction $w_i/\rho_k$ of its routing weight to it, so the weighted output $\gamma_{ij}e_j$ is replaced by $\gamma_{ij}e_i$, while the original weighted output $w_i e_i$ remains unchanged.

The resulting approximation error arises from using the retained experts to approximate the discarded experts.
Based on the weight redistribution in Proposition~\ref{prop:expert-weight-redistribution}, this error can be expressed as
\begin{equation}
\epsilon
=y-y'
=
\sum_{i\in\mathcal{S}_k}\sum_{j\in\mathcal{D}_k}
\gamma_{ij}(e_j-e_i).
\label{eq:expert-approximation-error}
\end{equation}

Eq.~(\ref{eq:expert-approximation-error}) highlights the role of expert similarity in determining the approximation error of Top-$k$ reduction.
When the similarity between retained and discarded experts is high, redistributing the discarded routing weights can compensate for the missing expert contributions.
When this similarity is low, however, redistribution may fail to provide effective compensation and instead increase the approximation error.
\insightbox{1}{The approximation error introduced by Top-$k$ reduction arises from using retained experts to approximate discarded experts and depends on expert similarity.}

%
%

\subsection{Timestep--Layer Heterogeneity of Local Error}
\label{sec:local-error-heterogeneity}

The preceding analysis characterizes the approximation error under a fixed Top-$k$ configuration.
However, the local errors introduced by different Top-$k$ configurations vary across denoising timesteps and network layers. To be specific, we compare the outputs under different Top-$k$ configurations with the default Top-$K$ output and measure local error using one minus the mean token-wise cosine similarity.

\begin{figure}[H]
\centering
\begin{subfigure}[b]{0.49\linewidth}
\centering
\includegraphics[width=\linewidth]{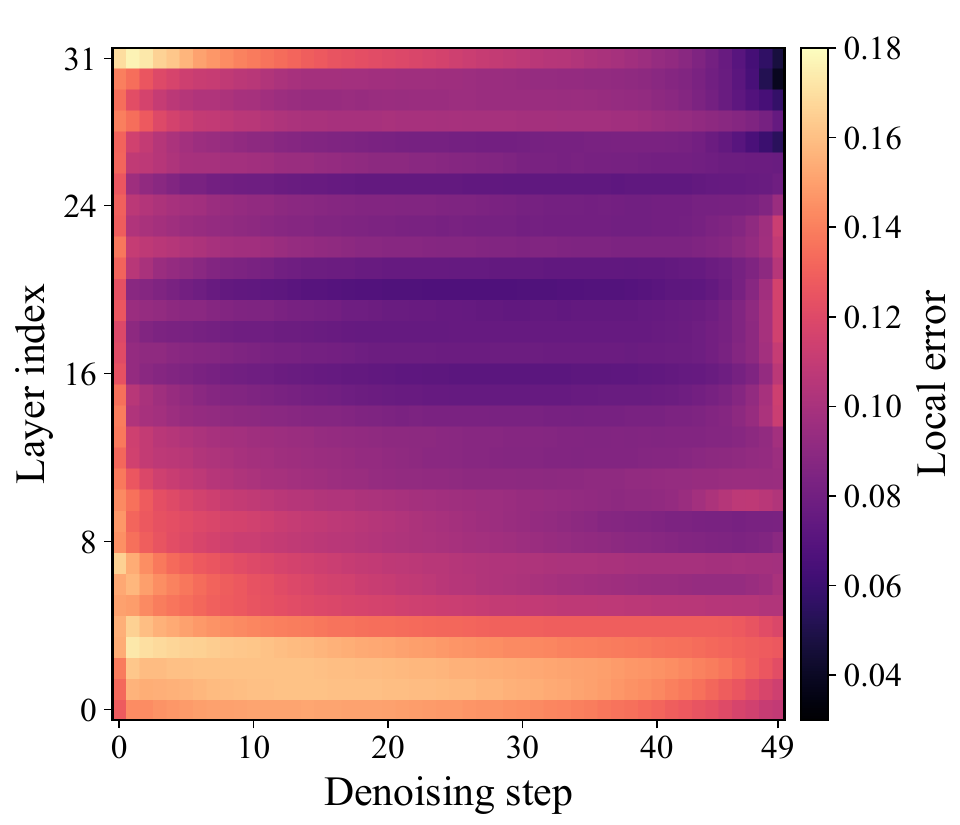}
\caption{Timestep--layer local error.}
\label{fig:local-error-heterogeneity}
\end{subfigure}\hfill
\begin{subfigure}[b]{0.49\linewidth}
\centering
\includegraphics[width=\linewidth]{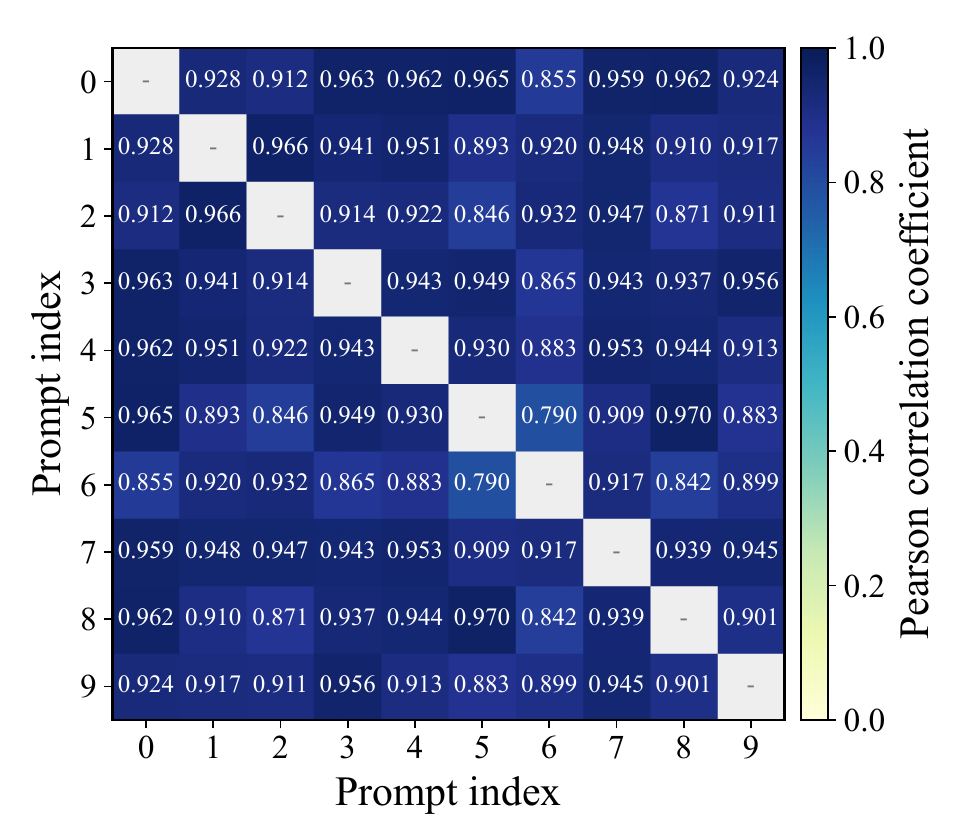}
\caption{Across-prompt error correlation.}
\label{fig:prompt-error-correlation}
\end{subfigure}
\caption{Local error analysis: (a) timestep--layer distribution of local error; (b) pairwise Pearson correlations between local error profiles across prompts.}
\label{fig:local-error-analysis}
\end{figure}

Figure~\ref{fig:local-error-analysis}\subref{fig:local-error-heterogeneity} shows the local error distribution across denoising timesteps and layers with Top-$4$, averaged over 10 prompts.
Even with the same number of active experts, local errors vary substantially across denoising timesteps and network layers, revealing pronounced timestep--layer heterogeneity.
A uniform Top-$k$ configuration therefore cannot fully accommodate these varying sensitivities, motivating joint expert budget allocation across timesteps and layers.

Besides, Figure~\ref{fig:local-error-analysis}\subref{fig:prompt-error-correlation} presents the pairwise Pearson correlations between local error profiles from different prompts.
The high mean correlation of 0.921 reveals consistent timestep--layer error patterns across different prompts, supporting the use of offline calibration to guide expert allocation.
\insightbox{2}{Local errors exhibit pronounced timestep--layer heterogeneity and maintain consistent patterns across prompts.}

\subsection{Timestep-Dependent Global Error}
\label{sec:trajectory-global-error}


Section~\ref{sec:local-error-heterogeneity} shows that local error varies across layers and timesteps, but local error is not equivalent to global error.
Local errors introduced by expert skipping propagate through subsequent denoising steps and may accumulate and be amplified, so errors of the same magnitude can affect the final output differently depending on when they occur.

Table~\ref{tab:trajectory-budget-ocr} in Appendix~\ref{app:timestep-expert-allocation} reports OCR scores~\citep{liu2026flow} for different allocation strategies under the same total expert budget.
Specifically, we divide the 50-step sampling process into the first 25 steps (F25) and the last 25 steps (L25) in denoising order, swap the expert budgets between the two halves while keeping the same average number of active experts, and use OCR scores to assess final generation quality.

Table~\ref{tab:trajectory-budget-ocr} highlights the timestep-dependent nature of global error, excessive expert reduction in either half degrades model performance.Too few activated experts in the early stage can disrupt global structure and introduce errors that accumulate and become amplified through subsequent steps, while too few in the later stage can impair detail recovery and final quality.


\insightbox{3}{Local errors at different timesteps have different impacts on global error, reflecting their timestep-dependent importance.}

\section{Methodology}
\label{sec:method}
\begin{figure}[H]
\centering
\includegraphics[width=\linewidth]{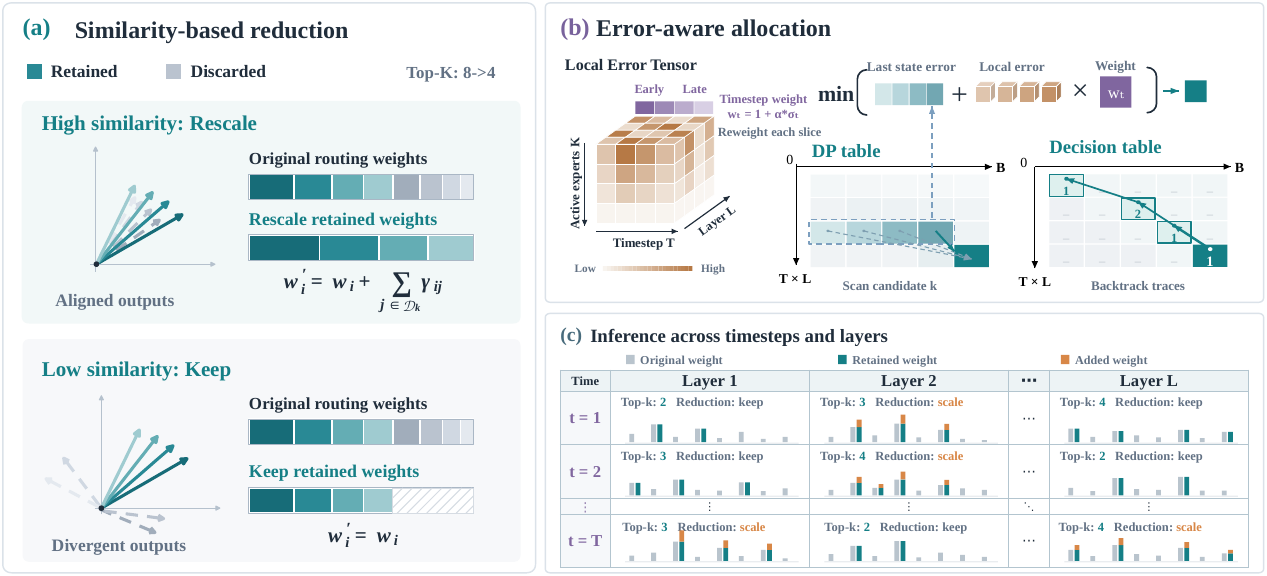}
\caption{Overview of ESEA. During calibration, ESEA determines routing weight adjustments using expert similarity and jointly allocates experts across timesteps and layers by minimizing timestep-weighted local errors with dynamic programming under a fixed total expert budget. During inference, the precomputed Top-$k$ configurations and weight adjustment decisions are applied directly.}
\label{fig:method-overview}
\end{figure}

Based on the error analysis in Section~\ref{sec:error-analysis}, we propose ESEA, a training-free and error-aware expert allocation method for efficient diffusion MoE inference, as illustrated in Figure~\ref{fig:method-overview}. ESEA first reduces approximation error through similarity-based Top-$k$ strategy, then constructs a local error tensor through offline calibration to capture timestep--layer heterogeneity. To account for the timestep-dependent impact of local errors on global error, it weights these errors according to the noise level. Finally, dynamic programming minimizes the weighted local error, jointly allocating experts across timesteps and layers under a fixed total expert budget.


\subsection{Similarity-Based Top-k Reduction}
\label{sec:similarity-based-topk-reduction}


As discussed in Section~\ref{sec:expert-approximation-error}, 
the approximation error introduced by Top-$k$ reduction arises from using retained experts to approximate discarded experts and depends on expert similarity.
We therefore propose a similarity-based Top-$k$ reduction strategy: when discarded experts are highly similar to the retained experts, we rescale the retained experts' routing weights; otherwise, we keep their original routing weights unchanged.This strategy is determined during offline calibration to avoid the overhead of computing token-wise expert similarities during inference.


We first run the model with its default Top-$K$ routing to collect calibration inputs at each denoising timestep $t$ and layer $l$, then feed these inputs to all experts to compute their pairwise expert similarities.
For a given Top-$k$ configuration, we use these precomputed similarities to average over all pairs of retained and discarded experts for each token, obtaining a token-level score.
Averaging these scores across tokens yields the layer-level similarity $s_{t,l}$ between retained and discarded experts.

\begin{equation}
y' =
\begin{cases}
\displaystyle
\sum_{i\in\mathcal{S}_k}
\left(w_i e_i + \sum_{j\in\mathcal{D}_k}\gamma_{ij}e_i\right),
& \text{if } s_{t,l}\geq 0.9, \\[8pt]
\displaystyle
\sum_{i\in\mathcal{S}_k} w_i e_i,
& \text{if } s_{t,l}<0.9.
\end{cases}
\label{eq:similarity-aware-output}
\end{equation}

Specifically, during inference, the precomputed similarity $s_{t,l}$ determines whether the retained experts' routing weights are rescaled to compensate for discarded contributions, as shown in Eq.~(\ref{eq:similarity-aware-output}).
This offline decision is shared by all tokens without introducing any additional online overhead, with an empirical comparison of the resulting approximation errors (Eq.~(\ref{eq:expert-approximation-error})) provided in Appendix~\ref{app:approximation-error-comparison}.

\subsection{Local Error-Aware Top-k Allocation}
\label{sec:local-error-aware-allocation}

Although the Top-$k$ reduction strategy has been determined, a uniform Top-$k$ configuration is not necessarily optimal.
As discussed in Insight~2 in Section~\ref{sec:local-error-heterogeneity}, local errors exhibit pronounced timestep--layer heterogeneity and maintain consistent patterns across prompts.
This motivates us to formulate it as a discrete allocation problem under a fixed total expert budget and solve it in two stages: local error tensor construction and optimal configuration search.

\begingroup
\setlength{\abovedisplayskip}{4pt plus 1pt minus 1pt}
\setlength{\belowdisplayskip}{4pt plus 1pt minus 1pt}
\setlength{\abovedisplayshortskip}{2pt plus 1pt minus 1pt}
\setlength{\belowdisplayshortskip}{3pt plus 1pt minus 1pt}
\textbf{Stage 1: Local Error Tensor Construction.}
To capture this timestep--layer heterogeneity, we construct a 3D local error tensor $\mathbf{S}\in\mathbb{R}^{T\times L\times K}$. Each entry $S[t,l,k]$ quantifies the local error introduced by reducing the default Top-$K$ configuration to Top-$k$ at timestep $t$ and layer $l$. Specifically, this error is computed as one minus the mean token-wise cosine similarity between the reduced output $y_{t,l}^{(k)}$ and the default output $y_{t,l}^{(K)}$:
\begin{equation}
S[t,l,k]
=1-\operatorname{cos}\left(y_{t,l}^{(k)},y_{t,l}^{(K)}\right).
\label{eq:local-error-tensor}
\end{equation}

Moreover, the consistent patterns across prompts observed in Insight~2 suggest that these error profiles are relatively stable across prompts. We therefore average local errors over the calibration set to obtain a reusable error tensor for online expert allocation.

\textbf{Stage 2: Optimal Configuration Search.}
Given the local error tensor $\mathbf{S}$, our objective is to identify an optimal Top-$k$ configuration across timesteps and layers under a fixed total expert budget. We define the total cost of a configuration as the sum of its local errors, leading to the following optimization problem:
\begin{equation}
k_{t,l}^{\star}
=
\underset{\{k_{t,l}\}}{\arg\min}
\sum_{t=0}^{T-1}\sum_{l=0}^{L-1}S[t,l,k_{t,l}],
\quad \text{s.t.}\quad
\sum_{t=0}^{T-1}\sum_{l=0}^{L-1}k_{t,l}=B,
\quad k_{t,l}\in\{1,\ldots,K\}.
\label{eq:optimal-configuration}
\end{equation}

where $B=\bar{k}TL$ denotes the total expert budget and $\bar{k}$ is the target average number of active experts.This discrete allocation problem can be efficiently solved using dynamic programming. We flatten the timestep--layer pairs into $N=TL$ decision units, with $(t_n,l_n)$ denoting the pair associated with the $n$-th unit. Let $D_n(b)$ denote the minimum cumulative local error achievable by the first $n$ units using exactly $b$ units of expert budget. Initializing $D_0(0)=0$ and all other states to $+\infty$, the recurrence is given by
\begin{equation}
D_n(b)=\min_{1\leq k\leq \min(K,b)}\left\{D_{n-1}(b-k)+S[t_n,l_n,k]\right\}.
\label{eq:dp-recurrence}
\end{equation}

For each state, we store the minimizing $k$ in a decision table $P_n(b)$. The minimum cumulative local error under the target budget is then given by $D_N(B)$. Starting from $b_N=B$, the optimal configuration is recovered by backtracking:
\begin{equation}
k_{t_n,l_n}^{\star}=P_n(b_n),
\qquad
b_{n-1}=b_n-k_{t_n,l_n}^{\star},
\quad n=N,\ldots,1.
\label{eq:dp-backtracking}
\end{equation}

With $N=TL$ decision units, the DP algorithm considers at most $K$ choices per budget state, yielding time complexity $O(NBK)$ and space complexity $O(NB)$. Given the consistent error patterns across prompts observed in Insight~2, we construct the error tensor using only a small set of calibration prompts, so the offline search incurs negligible overhead. 

\subsection{Global Error-Aware Timestep Weighting}
\label{sec:global-error-aware-weighting}

The preceding DP minimizes cumulative local error, but this objective does not directly reflect timestep-dependent impact on global error, as discussed in Insight~3 in Section~\ref{sec:trajectory-global-error}.
This motivates global error-aware timestep weighting to better reflect the differing contributions of local errors during dynamic programming.

To estimate this timestep-dependent importance, we use the noise level as a reference for the denoising stage.
Early high-noise timesteps primarily establish global structure, where perturbations can affect subsequent generation, while later low-noise timesteps refine local details.
For the flow-shifted sampling schedule, the noise level is defined as
\begin{equation}
u_t=1-\frac{t}{T},
\qquad
\sigma_t=\frac{s u_t}{1+(s-1)u_t},
\qquad t=0,\ldots,T-1,
\label{eq:flow-shifted-noise-level}
\end{equation}
where $s$ is the flow-shift parameter.We then define the timestep weight and the weighted local error tensor as
\begin{equation}
w_t=1+\alpha\sigma_t,
\qquad
\widetilde{S}[t,l,k]=w_t S[t,l,k],
\label{eq:timestep-weight}
\end{equation}

Since $\sigma_t$ decreases throughout denoising, earlier timesteps receive larger weights, encouraging the allocation to preserve more expert computation during global structure formation. However, overly aggressive weighting can leave insufficient budget for later timesteps, impairing detail reconstruction or causing blurred outputs. The parameter $\alpha$ therefore controls the balance between early structure preservation and late detail refinement.

In practice, we select the optimal $\alpha$ through grid search over $[0,1]$ with a step size of $0.2$. By replacing $\mathbf{S}$ with $\widetilde{\mathbf{S}}$ in Eq.~\ref{eq:timestep-weight}, we extend the objective from minimizing cumulative local error to minimizing cumulative timestep-dependent global error under the original expert budget:
\begin{equation}
k_{t,l}^{\star}
=
\underset{\{k_{t,l}\}}{\arg\min}
\sum_{t=0}^{T-1}\sum_{l=0}^{L-1}w_t S[t,l,k_{t,l}],
\quad \text{s.t.}\quad
\sum_{t=0}^{T-1}\sum_{l=0}^{L-1}k_{t,l}=B,
\quad k_{t,l}\in\{1,\ldots,K\}.
\label{eq:weighted-expert-allocation}
\end{equation}
\par\endgroup

\section{Experiment}
\label{sec:experiment}

\subsection{Experimental Setup}
\label{sec:experimental-setup}

\textbf{Base Models and Baselines.} To evaluate the efficacy of our proposed method, we select two representative diffusion MoE models: HunyuanImage-3.0-80B-A13B~\citep{team2025hunyuanimage} for image generation and LingBot-Video-30B-A3B~\citep{ma2026scaling} for video generation.
Additionally, as there are no expert skipping baselines for diffusion MoE models, we select two representative methods: NAEE~\citep{lu2024not} and MoDES~\citep{huang2026modes}, both of which we re-implement and adapt strictly following their official settings.

\textbf{Evaluation Metrics.} 
For HunyuanImage-3.0-Instruct and its distilled variant, we use ImageReward~\citep{xu2023imagereward} to assess overall generation quality, alongside PickScore~\citep{kirstain2023pick} and AestheticScore~\citep{verb2024aesthetic} for human preference and visual aesthetics, respectively.
We further evaluate text rendering accuracy with OCR~\citep{liu2026flow}, compositional instruction following with GenEval~\citep{ghosh2023geneval}, and text--image alignment with CLIPScore~\citep{hessel2021clipscore}.
For LingBot-Video, we follow the official evaluation protocol on RBench~\citep{deng2026rethinking}, covering five task categories and four embodiment categories.
Speedup is computed relative to the no expert skipping baselines.

\textbf{Implementation Details.} All experiments are implemented in PyTorch and conducted on NVIDIA A800 40GB GPUs for HunyuanImage-3.0 and NVIDIA H20 80GB GPUs for LingBot-Video.
Further details regarding evaluation and method implementation are provided in Appendix~\ref{app:experimental-details}.

\subsection{Main Result}
\label{sec:main-result}
As shown in Tables~\ref{tab:main-image-results} and~\ref{tab:main-video-results}, ESEA consistently outperforms uniform Top-$4$, NAEE, and MoDES in overall generation quality under the same average expert budget. On HunyuanImage-3.0-Instruct, ESEA achieves an ImageReward of 1.0973, compared with 0.9785 for uniform Top-$4$, 0.7874 for NAEE, and 0.9221 for MoDES, remaining close to the original Top-$8$ model (1.1114). Beyond overall generation quality, ESEA achieves the highest scores among the compared expert skipping methods on all evaluation metrics. Similar gains are observed on the distilled variant, where ESEA improves ImageReward from 0.8438 to 0.9887 over uniform Top-$4$. Notably, both NAEE and MoDES obtain lower overall scores than uniform Top-$4$, suggesting that dynamic expert skipping without accounting for timestep-dependent sensitivity can further degrade generation quality.

On LingBot-Video, ESEA achieves an average RBench score of 0.835, outperforming uniform Top-$4$ (0.805), NAEE (0.794), and MoDES (0.813). ESEA obtains the best or tied-best scores among the compared expert skipping methods in eight of the nine evaluated categories. 
Additional results under different Top-$k$ configurations and qualitative comparisons are provided in Appendix~\ref{app:additional-expert-budgets} and Appendix~\ref{app:qualitative-comparisons}, respectively.

\begin{table}[H]
\centering
\caption{Quantitative evaluation of generation quality and inference efficiency on HunyuanImage-3.0-Instruct and its distilled variant. All reduced methods target an average Top-$k$ of 4.}
\label{tab:main-image-results}
\footnotesize
\setlength{\tabcolsep}{2pt}
\begin{tabularx}{\linewidth}{l|>{\hsize=1.25\hsize\linewidth=\hsize\centering\arraybackslash}X|>{\hsize=1.03\hsize\linewidth=\hsize\centering\arraybackslash}X>{\hsize=1.35\hsize\linewidth=\hsize\centering\arraybackslash}X|>{\hsize=0.65\hsize\linewidth=\hsize\centering\arraybackslash}X>{\hsize=0.85\hsize\linewidth=\hsize\centering\arraybackslash}X>{\hsize=0.72\hsize\linewidth=\hsize\centering\arraybackslash}X|>{\hsize=1.15\hsize\linewidth=\hsize\centering\arraybackslash}X}
\toprule
\multicolumn{1}{c|}{Method} & Overall & \multicolumn{2}{c|}{Image Quality} & \multicolumn{3}{c|}{Text Alignment} & Acceleration \\
\cmidrule(lr){2-2}\cmidrule(lr){3-4}\cmidrule(lr){5-7}\cmidrule(lr){8-8}
 & ImageReward$\uparrow$ & PickScore$\uparrow$ & AestheticScore$\uparrow$ & OCR$\uparrow$ & GenEval$\uparrow$ & CLIP$\uparrow$ & Speedup$\uparrow$ \\
\midrule
\multicolumn{8}{c}{\textbf{(a) HunyuanImage-3.0-Instruct (80B-A13B)}} \\
\midrule
Top-$8$ & 1.1114 & 22.3786 & 5.6999 & 0.7932 & 0.8619 & 29.5973 & $1.0000\times$ \\
\midrule
Top-$4$ & 0.9785 & 22.1368 & 5.6027 & 0.6379 & 0.8320 & 28.4033 & $\boldsymbol{1.8000\times}$ \\
NAEE & 0.7874 & 21.7987 & 5.0172 & 0.5152 & 0.7709 & 27.2839 & $1.7888\times$ \\
MoDES & 0.9221 & 20.9976 & 4.9846 & 0.4415 & 0.8121 & 29.0843 & $1.7817\times$ \\
\midrule
\rowcolor{InsightBlue!12}
\textbf{ESEA} & \textbf{1.0973} & \textbf{22.3053} & \textbf{5.6267} & \textbf{0.7440} & \textbf{0.8670} & \textbf{29.1966} & $1.7962\times$ \\
\midrule
\multicolumn{8}{c}{\textbf{(b) HunyuanImage-3.0-Instruct-Distil (80B-A13B)}} \\
\midrule
Top-$8$ & 1.0248 & 22.1205 & 5.6707 & 0.7450 & 0.8889 & 28.5997 & $1.0000\times$ \\
\midrule
Top-$4$ & 0.8438 & 21.8349 & 5.5565 & 0.5632 & 0.8197 & 27.5620 & $\boldsymbol{1.8000\times}$ \\
NAEE & 0.5071 & 21.3650 & 5.5051 & 0.3847 & 0.7221 & 26.2040 & $1.7888\times$ \\
MoDES & 0.8293 & 21.1046 & 5.2378 & 0.4906 & 0.8465 & 27.8956 & $1.7597\times$ \\
\midrule
\rowcolor{InsightBlue!12}
\textbf{ESEA} & \textbf{0.9887} & \textbf{21.9920} & \textbf{5.5938} & \textbf{0.7013} & \textbf{0.8748} & \textbf{28.3670} & $1.7824\times$ \\
\bottomrule
\end{tabularx}
\end{table}

\begin{table}[H]
\centering
\caption{RBench performance and inference efficiency on LingBot-Video (30B-A3B). All reduced methods use an average Top-$k$ of 4.}
\label{tab:main-video-results}
\footnotesize
\setlength{\tabcolsep}{1.5pt}
\begin{tabularx}{\linewidth}{l|>{\centering\arraybackslash}X|*{4}{>{\centering\arraybackslash}X}>{\hsize=1.1\hsize\linewidth=\hsize\centering\arraybackslash}X|*{3}{>{\centering\arraybackslash}X}>{\hsize=0.9\hsize\linewidth=\hsize\centering\arraybackslash}X|c}
\toprule
\multicolumn{12}{c}{\textbf{LingBot-Video (30B-A3B)}} \\
\midrule
\multicolumn{1}{c|}{Method} & Overall & \multicolumn{5}{c|}{Task Scores$\uparrow$} & \multicolumn{4}{c|}{Embodiment Scores$\uparrow$} & Acceleration \\
\cmidrule(lr){2-2}\cmidrule(lr){3-7}\cmidrule(lr){8-11}\cmidrule(lr){12-12}
& Avg$\uparrow$ & Manip. & Spatial & Multi. & Long. & Reason. & Single. & Dual. & Quad. & Hum. & Speedup$\uparrow$ \\
\midrule
Top-$8$ & 0.829 & 0.928 & 0.860 & 0.922 & 0.947 & 0.897 & 0.694 & 0.723 & 0.734 & 0.760 &   $1.0000\times$\\
\midrule
Top-$4$ & 0.805 & 0.898 & 0.860 & \textbf{0.898} & 0.908 & 0.912 & 0.678 & 0.671 & 0.681 & 0.743 &   $\boldsymbol{1.8000\times}$\\
NAEE & 0.794 & 0.900 & 0.840 & 0.825 & 0.938 & 0.880 & 0.676 & 0.665 & 0.670 & 0.750 &   $1.6452\times$\\
MoDES & 0.813 & 0.912 & 0.860 & 0.810 & 0.956 & 0.910 & 0.682 & 0.684 & 0.742 & 0.762 &   $1.7173\times$\\
\midrule
\rowcolor{InsightBlue!12}
\textbf{ESEA} & \textbf{0.835} & \textbf{0.940} & \textbf{0.860} & 0.880 & \textbf{0.959} & \textbf{0.945} & \textbf{0.692} & \textbf{0.702} & \textbf{0.748} & \textbf{0.786} &  $1.7832\times$\\
\bottomrule
\end{tabularx}
\end{table}

\subsection{Ablation Study}
\label{sec:ablation-study}
We ablate three components of ESEA: similarity-based Top-$k$ reduction, local error-aware Top-$k$ allocation, and global error-aware timestep weighting. As shown in Tables~\ref{tab:ablation-image-results} and~\ref{tab:ablation-video-results} in Appendix~\ref{app:component-ablation}, similarity-based reduction improves overall generation quality over uniform Top-$4$ by adjusting routing weights according to expert similarity. Local error-aware allocation yields additional gains by capturing timestep--layer heterogeneity in local errors, while global error-aware weighting further improves performance by accounting for their timestep-dependent impacts on global error. Across HunyuanImage-3.0-Instruct, its distilled variant, and LingBot-Video, the complete method achieves the best overall performance among all ablation configurations.


\section{Conclusion}
We present ESEA, a training-free and error-aware expert allocation method for diffusion MoE inference. Our analysis reveals the role of expert similarity in approximation error, timestep--layer heterogeneity in local errors, and the timestep-dependent impacts of local errors on global error. Based on these insights, ESEA combines similarity-based routing weight adjustment with dynamic programming to minimize timestep-weighted local errors, jointly allocating experts across timesteps and layers under a fixed total expert budget. The resulting configurations are determined offline and retrieved during inference with negligible overhead. Experiments on HunyuanImage-3.0-Instruct, its distilled variant, and LingBot-Video demonstrate that ESEA achieves the best overall generation quality among the compared expert skipping methods under the same expert budget.

\subsection*{AI use statement}

AI tools were used to aid in the writing and polishing of the manuscript.
Specifically, we used AI to assist in refining the language, improving readability, and ensuring clarity in various sections of the paper.
The model helped with tasks such as sentence rephrasing, grammar checking, and enhancing the overall flow of the text.

It is important to note that AI was not involved in the ideation, research methodology, or experimental design.
All research concepts, ideas, and analyses were developed and conducted by the authors.
The contributions of AI were solely focused on improving the linguistic quality of the paper, with no involvement in the scientific content or data analysis.

The authors take full responsibility for the content of the manuscript, including any text generated or polished by AI.
We have ensured that the AI-generated text adheres to ethical guidelines and does not contribute to plagiarism or scientific misconduct.

\subsection*{Ethics statement}

This work adheres to the ICLR Code of Ethics.
In this study, no human subjects were involved, and no animal experiments were conducted.
We have taken care to avoid any biases or discriminatory outcomes in our research process.
No personally identifiable information was used, and no experiments were conducted that could raise privacy or security concerns.
We are committed to maintaining transparency and integrity throughout the research process.

\subsection*{Reproducibility statement}

We have made every effort to ensure that the results presented in this paper are reproducible.
All datasets used in our experiments, as described in Appendix~\ref{app:evaluation-details}, are publicly available, ensuring consistent and reproducible evaluation results.
The experimental setup, including method implementation, baseline adaptations, evaluation protocols, and generation settings, is described in detail in Section~\ref{sec:experimental-setup} and Appendix~\ref{app:experimental-details}.
We believe these measures will enable other researchers to reproduce our work and further advance the field.



\bibliography{iclr2027_conference}
\bibliographystyle{iclr2027_conference}

\appendix
\section{Detailed Proof of Proposition~\ref{prop:expert-weight-redistribution}}
\label{app:expert-weight-redistribution-proof}


\paragraph{Restatement of Proposition~\ref{prop:expert-weight-redistribution} (Expert Weight Redistribution).}
\begingroup
\itshape
Let $\rho_k=\sum_{i\in\mathcal{S}_k}w_i$ and $\delta_k=\sum_{j\in\mathcal{D}_k}w_j$ be the total routing weights of the retained and discarded experts, respectively.
With $\gamma_{ij}=w_iw_j/\rho_k$ denoting the routing weight transferred from discarded expert $j$ to retained expert $i$, the output $y'$ under reduced Top-$k$ routing can be expressed as
\begin{equation}
y'
=\sum_{i\in\mathcal{S}_k}
\left(w_i e_i+\sum_{j\in\mathcal{D}_k}\gamma_{ij}e_i\right).
\end{equation}
\par\endgroup

\begin{proof}
Consider the standard MoE forward pass, whose output can be decomposed into the contributions of the retained and discarded experts:
\begin{equation}
y
=\sum_{i\in\mathcal{S}_k}w_i e_i
+\sum_{j\in\mathcal{D}_k}w_j e_j.
\end{equation}
Reducing Top-$K$ to Top-$k$ proceeds in two stages: truncating the discarded experts and normalizing the routing weights of the retained experts.
Truncation removes the contributions of the experts in $\mathcal{D}_k$ while leaving the retained weights unchanged, giving
\begin{equation}
y_{\mathrm{trunc}}
=y-\sum_{j\in\mathcal{D}_k}w_j e_j
=\sum_{i\in\mathcal{S}_k}w_i e_i.
\end{equation}
Normalization then increases each retained weight by a factor of $1/\rho_k$, so that the retained weights sum to one:
\begin{equation}
\widetilde w_i
=\frac{w_i}{\rho_k}
=w_i+\frac{\delta_k}{\rho_k}w_i,
\qquad i\in\mathcal{S}_k.
\label{eq:appendix-proposition-weight-rescaling}
\end{equation}
The total increase is $\sum_{i\in\mathcal{S}_k}(\widetilde w_i-w_i)=\delta_k$, showing that normalization redistributes the discarded routing weight to the retained experts in proportion to their original weights.
For a retained expert $i\in\mathcal{S}_k$, expanding $\delta_k$ over the discarded experts reveals where its additional routing weight comes from:
\begin{equation}
\widetilde w_i-w_i
=\frac{\delta_k}{\rho_k}w_i
=\sum_{j\in\mathcal{D}_k}\frac{w_iw_j}{\rho_k}
=\sum_{j\in\mathcal{D}_k}\gamma_{ij},
\qquad
\gamma_{ij}=\frac{w_iw_j}{\rho_k}.
\label{eq:appendix-proposition-incoming-weights}
\end{equation}
Under this decomposition, each discarded expert $j$ transfers a fraction $w_i/\rho_k$ of its routing weight to expert $i$.
The entire weight of expert $j$ is distributed among the retained experts, since
\begin{equation}
\sum_{i\in\mathcal{S}_k}\gamma_{ij}
=\frac{w_j}{\rho_k}\sum_{i\in\mathcal{S}_k}w_i
=w_j.
\label{eq:appendix-proposition-weight-conservation}
\end{equation}
The transferred weight $\gamma_{ij}$ originally multiplies the discarded expert's output $e_j$, but after redistribution it multiplies the retained expert's output $e_i$, replacing the weighted output $\gamma_{ij}e_j$ with $\gamma_{ij}e_i$.
Substituting $\widetilde w_i=w_i+\sum_{j\in\mathcal{D}_k}\gamma_{ij}$ into the reduced MoE output yields the expression in Eq.~(\ref{eq:expert-redistribution-output}):
\begin{equation}
\begin{aligned}
y'
&=\sum_{i\in\mathcal{S}_k}\widetilde w_i e_i\\
&=\sum_{i\in\mathcal{S}_k}
\left(w_i e_i+\sum_{j\in\mathcal{D}_k}\gamma_{ij}e_i\right).
\end{aligned}
\label{eq:appendix-proposition-redistributed-output}
\end{equation}
\end{proof}

\section{Timestep-Dependent Expert Allocation}
\label{app:timestep-expert-allocation}

F25 and L25 denote the first 25 and last 25 steps of the 50-step denoising process, respectively.A moderate shift toward early computation improves OCR scores from 0.6379 with uniform Top-$4$ to 0.7274 with F25 Top-$5$/L25 Top-$3$, whereas further shifting to F25 Top-$7$/L25 Top-$1$ lowers the score to 0.1927.

\begin{table}[H]
\centering
\caption{OCR scores for different expert allocation configurations at an average Top-$k$ of 4.}
\label{tab:trajectory-budget-ocr}
\begin{tabular}{cccc}
\hline
F25 Top-$k$ & L25 Top-$k$ & Avg. Top-$k$ & OCR score $\uparrow$ \\
\hline
8 & 8 & 8 & 0.7932 \\
\hline
1 & 7 & 4 & 0.0532 \\
2 & 6 & 4 & 0.1928 \\
3 & 5 & 4 & 0.5118 \\
4 & 4 & 4 & 0.6379 \\
\textbf{5} & \textbf{3} & \textbf{4} & \textbf{0.7274} \\
6 & 2 & 4 & 0.7227 \\
7 & 1 & 4 & 0.1927 \\
\hline
\end{tabular}
\end{table}

\section{Empirical Comparison of Approximation Errors}
\label{app:approximation-error-comparison}
\begin{figure}[H]
\centering
\includegraphics[width=\linewidth]{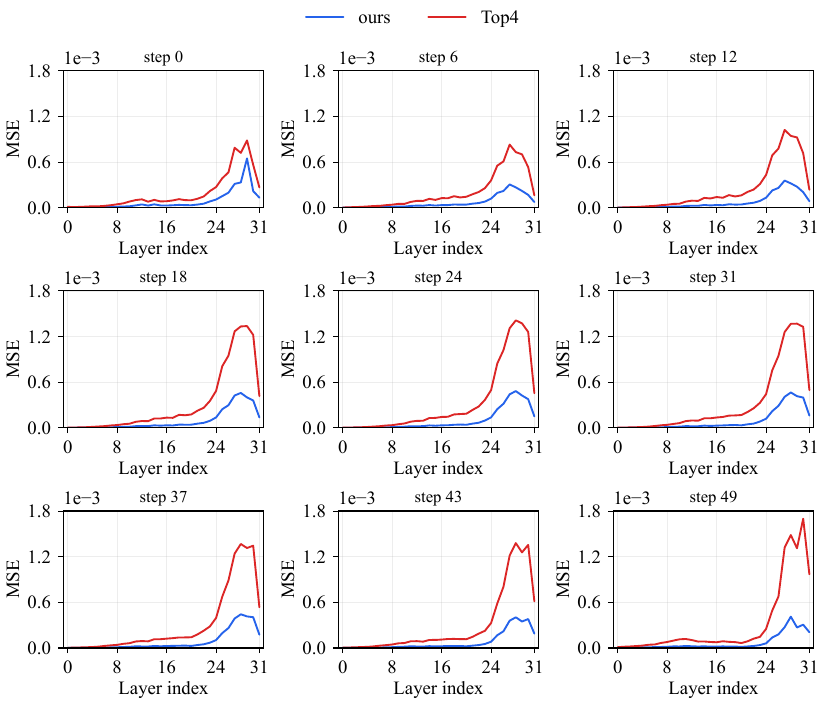}
\caption{Layer-wise approximation errors of our similarity-based Top-$4$ reduction and standard Top-$4$ routing. MSE is measured relative to the original output without expert skipping.}
\label{fig:approximation-error-comparison}
\end{figure}

We compare the layer-wise approximation errors of similarity-based Top-$4$ reduction and standard Top-$4$ routing at different denoising steps, measuring MSE relative to the original output without expert skipping.
As shown in Figure~\ref{fig:approximation-error-comparison}, similarity-based reduction consistently lowers approximation errors across the sampled timesteps.
The reductions are particularly pronounced in deeper layers, where standard Top-$4$ routing exhibits large error peaks.
These results demonstrate that adjusting routing weights according to expert similarity effectively reduces approximation error at the same number of active experts.

\section{Additional Results at Different Expert Budgets}
\label{app:additional-expert-budgets}

We further evaluate ESEA at average Top-$k$ budgets of 5 and 3 on HunyuanImage-3.0-Instruct, its distilled variant, and LingBot-Video.
As shown in Tables~\ref{tab:additional-image-budgets} and~\ref{tab:additional-video-budgets}, ESEA consistently improves overall generation quality over uniform Top-$k$ routing under the same expert budget.
The gains are more pronounced at the Top-$3$ budget, where ESEA increases ImageReward from 0.4660 to 0.9314 on the distilled image model and the average RBench score from 0.755 to 0.835 on LingBot-Video.  
These results demonstrate the effectiveness of ESEA across different expert budgets, particularly in preserving generation quality under more aggressive expert skipping.


\begin{table}[H]
\centering
\caption{Generation quality at average Top-$k$ budgets of 5 and 3 on HunyuanImage-3.0-Instruct and its distilled variant, with the original Top-$8$ model as a reference.}
\label{tab:additional-image-budgets}
\footnotesize
\setlength{\tabcolsep}{2pt}
\begin{tabularx}{\linewidth}{l|>{\hsize=1.25\hsize\linewidth=\hsize\centering\arraybackslash}X|>{\hsize=1.0\hsize\linewidth=\hsize\centering\arraybackslash}X>{\hsize=1.35\hsize\linewidth=\hsize\centering\arraybackslash}X|>{\hsize=0.7\hsize\linewidth=\hsize\centering\arraybackslash}X>{\hsize=1.0\hsize\linewidth=\hsize\centering\arraybackslash}X>{\hsize=0.7\hsize\linewidth=\hsize\centering\arraybackslash}X}
\toprule
\multicolumn{1}{c|}{Method} & Overall & \multicolumn{2}{c|}{Image Quality} & \multicolumn{3}{c}{Text Alignment} \\
\cmidrule(lr){2-2}\cmidrule(lr){3-4}\cmidrule(lr){5-7}
 & ImageReward$\uparrow$ & PickScore$\uparrow$ & AestheticScore$\uparrow$ & OCR$\uparrow$ & GenEval$\uparrow$ & CLIP$\uparrow$ \\
\midrule
\multicolumn{7}{c}{\textbf{(a) HunyuanImage-3.0-Instruct (80B-A13B)}} \\
\midrule
Top-$8$ & 1.1114 & 22.3786 & 5.6999 & 0.7932 & 0.8619 & 29.5973 \\
Top-$5$ & 1.0298 & 22.2754 & 5.6119 & 0.7152 & 0.8603 & 28.8836 \\
\rowcolor{InsightBlue!12}
\textbf{ESEA (Top-$\mathbf{5}$)} & \textbf{1.1263} & \textbf{22.3704} & \textbf{5.6551} & \textbf{0.7761} & \textbf{0.8736} & \textbf{29.4374} \\
Top-$3$ & 0.8555 & 21.8807 & \textbf{5.5327} & 0.4765 & 0.7724 & 27.5820 \\
\rowcolor{InsightBlue!12}
\textbf{ESEA (Top-$\mathbf{3}$)} & \textbf{1.1048} & \textbf{22.0807} & 5.5045 & \textbf{0.7024} & \textbf{0.8450} & \textbf{29.1341} \\
\midrule
\multicolumn{7}{c}{\textbf{(b) HunyuanImage-3.0-Instruct-Distil (80B-A13B)}} \\
\midrule
Top-$8$ & 1.0248 & 22.1205 & 5.6707 & 0.7450 & 0.8889 & 28.5997 \\
Top-$5$ & 0.9249 & 22.0005 & 5.6084 & 0.6915 & 0.8556 & 28.3148 \\
\rowcolor{InsightBlue!12}
\textbf{ESEA (Top-$\mathbf{5}$)} & \textbf{0.9936} & \textbf{22.0789} & \textbf{5.6526} & \textbf{0.7339} & \textbf{0.8887} & \textbf{28.3989} \\
Top-$3$ & 0.4660 & 21.2465 & 5.3951 & 0.2851 & 0.7382 & 25.9148 \\
\rowcolor{InsightBlue!12}
\textbf{ESEA (Top-$\mathbf{3}$)} & \textbf{0.9314} & \textbf{21.7520} & \textbf{5.5323} & \textbf{0.6531} & \textbf{0.8622} & \textbf{27.8377} \\
\bottomrule
\end{tabularx}
\end{table}

\begin{table}[H]
\centering
\caption{RBench performance at average Top-$k$ budgets of 5 and 3 on LingBot-Video (30B-A3B), with the original Top-$8$ model as a reference.}
\label{tab:additional-video-budgets}
\footnotesize
\setlength{\tabcolsep}{1pt}
\begin{tabularx}{\linewidth}{l|>{\centering\arraybackslash}X|*{5}{>{\centering\arraybackslash}X}|*{4}{>{\centering\arraybackslash}X}}
\toprule
\multicolumn{11}{c}{\textbf{LingBot-Video (30B-A3B)}} \\
\midrule
\multicolumn{1}{c|}{Method} & Overall & \multicolumn{5}{c|}{Task Scores$\uparrow$} & \multicolumn{4}{c}{Embodiment Scores$\uparrow$} \\
\cmidrule(lr){2-2}\cmidrule(lr){3-7}\cmidrule(lr){8-11}
& Avg$\uparrow$ & Manip. & Spatial & Multi. & Long. & Reason. & Single. & Dual. & Quad. & Hum. \\
\midrule
Top-$8$ & 0.829 & 0.928 & 0.860 & 0.922 & 0.947 & 0.897 & 0.694 & 0.723 & 0.734 & 0.760 \\
Top-$5$ & 0.801 & 0.910 & \textbf{0.820} & 0.855 & 0.935 & 0.872 & 0.672 & 0.684 & 0.710 & 0.755 \\
\rowcolor{InsightBlue!12}
\textbf{ESEA (Top-$\mathbf{5}$)} & \textbf{0.818} & \textbf{0.920} & 0.780 & \textbf{0.890} & \textbf{0.944} & \textbf{0.912} & \textbf{0.698} & \textbf{0.692} & \textbf{0.754} & \textbf{0.773} \\
Top-$3$ & 0.755 & 0.855 & 0.830 & 0.760 & 0.893 & 0.891 & 0.637 & 0.626 & 0.610 & 0.696 \\
\rowcolor{InsightBlue!12}
\textbf{ESEA (Top-$\mathbf{3}$)} & \textbf{0.835} & \textbf{0.928} & \textbf{0.880} & \textbf{0.908} & \textbf{0.934} & \textbf{0.930} & \textbf{0.720} & \textbf{0.688} & \textbf{0.774} & \textbf{0.753} \\
\bottomrule
\end{tabularx}
\end{table}

\section{Additional Experimental Details}
\label{app:experimental-details}

\subsection{Evaluation}
\label{app:evaluation-details}
For HunyuanImage-3.0, we evaluate both the Instruct and Instruct-Distil variants at a resolution of $1024\times1024$ with 50 and 8 NFEs, respectively. 
Each evaluation uses a subset of 300 test prompts sampled randomly from the corresponding test split.
We use the official GenEval test prompts~\citep{ghosh2023geneval} to evaluate compositional instruction following, while ImageReward, AestheticScore, and CLIPScore are evaluated on DrawBench~\citep{saharia2022photorealistic}.
PickScore is evaluated on Pick-a-Pic~\citep{kirstain2023pick}, and OCR uses the visual text rendering test split provided by Flow-GRPO~\citep{liu2026flow}.

For LingBot-Video, we generate 121-frame videos at 24 FPS with 40 NFEs, with the output resolution matching that of the corresponding RBench reference images.
And we evaluate on the complete RBench benchmark~\citep{deng2026rethinking}, which contains 650 text--image prompts.
The task-oriented split consists of 250 prompts, with 50 prompts for each of five categories: common manipulation, spatial relationship, multi-entity collaboration, long-horizon planning, and visual reasoning.
The embodiment-oriented split consists of 400 prompts, with 100 prompts for each of four robot morphologies: single-arm, dual-arm, humanoid, and quadruped robots.
In addition, we use Qwen3-VL-30B-A3B-Instruct~\citep{bai2025qwen3} as the VLM evaluator.

\subsection{Method Implementation}
\label{app:method-implementation}
We strictly follow the expert-skipping rules in the original papers, adopting the generalized Top-$K$ NAEE implementation described in MoDES~\citep{huang2026modes}, with details below.

\textbf{MoDES.} Following MoDES~\citep{huang2026modes}, we combine an offline-calibrated layer importance factor $\alpha^{(l)}$ with each token's routing weight $w_m^{(l)}$, skipping experts for which $\alpha^{(l)}w_m^{(l)}<\tau$.
In our diffusion adaptation, routed tokens are treated as a single modality, reducing the original dual-modality threshold search to a one-dimensional search over a single threshold $\tau$ shared across layers and denoising timesteps.
Layer importance is calibrated by disabling one layer's routed experts throughout the denoising trajectory, measuring the MSE between the resulting final latent and the unmodified model's final latent, and normalizing these errors across layers.

\textbf{NAEE.} We implement NAEE~\citep{lu2024not} using the Top-$K$ extension in MoDES~\citep{huang2026modes}.
For each token, the original $K$ selected experts are ordered by decreasing routing weight, denoted by $w_{(1)}^{(l)}\geq\cdots\geq w_{(K)}^{(l)}$, and the largest low-weight suffix satisfying
\begin{equation}
\sum_{r=i}^{K}w_{(r)}^{(l)}
<\beta^{(l)}\sum_{r=1}^{K}w_{(r)}^{(l)},
\qquad 2\leq i\leq K,
\label{eq:naee-skipping-rule}
\end{equation}
is skipped, ensuring that at least one expert remains active; if no suffix satisfies this condition, all $K$ experts are retained.
The layer-specific thresholds $\beta^{(l)}$ are optimized by genetic search under the target average expert budget constraint.

\section{Component Ablation Results}
\label{app:component-ablation}

\begin{table}[H]
\centering
\caption{Component ablation on HunyuanImage-3.0-Instruct and its distilled variant at an average Top-$k$ of 4. Components: (1) Similarity-Based Top-$k$ Reduction; (2) Local Error-Aware Top-$k$ Allocation; (3) Global Error-Aware Timestep Weighting. Disabling all components gives uniform Top-$4$, while enabling all three gives ESEA.}
\label{tab:ablation-image-results}
\footnotesize
\setlength{\tabcolsep}{2pt}
\begin{tabularx}{\linewidth}{ccc|>{\hsize=1.25\hsize\linewidth=\hsize\centering\arraybackslash}X|>{\hsize=1.0\hsize\linewidth=\hsize\centering\arraybackslash}X>{\hsize=1.35\hsize\linewidth=\hsize\centering\arraybackslash}X|>{\hsize=0.7\hsize\linewidth=\hsize\centering\arraybackslash}X>{\hsize=1.0\hsize\linewidth=\hsize\centering\arraybackslash}X>{\hsize=0.7\hsize\linewidth=\hsize\centering\arraybackslash}X}
\toprule
\multicolumn{3}{c|}{Components} & Overall & \multicolumn{2}{c|}{Image Quality} & \multicolumn{3}{c}{Text Alignment} \\
\cmidrule(lr){1-3}\cmidrule(lr){4-4}\cmidrule(lr){5-6}\cmidrule(lr){7-9}
(1) & (2) & (3) & ImageReward$\uparrow$ & PickScore$\uparrow$ & AestheticScore$\uparrow$ & OCR$\uparrow$ & GenEval$\uparrow$ & CLIP$\uparrow$ \\
\midrule
\multicolumn{9}{c}{\textbf{(a) HunyuanImage-3.0-Instruct (80B-A13B)}} \\
\midrule
-- & -- & -- & 0.9785 & 22.1368 & 5.6027 & 0.6379 & 0.8320 & 28.4033 \\
\midrule
$\checkmark$ & -- & -- & 1.0532 & 22.2324 & 5.6063 & 0.7225 & 0.8466 & 29.0580 \\
$\checkmark$ & $\checkmark$ & -- & 1.0895 & 22.2971 & 5.6167 & 0.7411 & \textbf{0.8702} & 29.1875 \\
\midrule
$\checkmark$ & $\checkmark$ & $\checkmark$ & \textbf{1.0973} & \textbf{22.3053} & \textbf{5.6267} & \textbf{0.7440} & 0.8670 & \textbf{29.1966} \\
\midrule
\multicolumn{9}{c}{\textbf{(b) HunyuanImage-3.0-Instruct-Distil (80B-A13B)}} \\
\midrule
-- & -- & -- & 0.8438 & 21.8349 & 5.5565 & 0.5632 & 0.8197 & 27.5620 \\
\midrule
$\checkmark$ & -- & -- & 0.9426 & 21.9229 & 5.5358 & \textbf{0.7151} & 0.8716 & 28.1332 \\
$\checkmark$ & $\checkmark$ & -- & 0.9693 & 21.9800 & 5.5348 & 0.6982 & 0.8614 & 28.2034 \\
\midrule
$\checkmark$ & $\checkmark$ & $\checkmark$ & \textbf{0.9887} & \textbf{21.9920} & \textbf{5.5938} & 0.7013 & \textbf{0.8748} & \textbf{28.3670} \\
\bottomrule
\end{tabularx}
\end{table}

\begin{table}[H]
\centering
\caption{Component ablation on LingBot-Video (30B-A3B) at an average Top-$k$ of 4. Components (1)--(3) follow Table~\ref{tab:ablation-image-results}.}
\label{tab:ablation-video-results}
\footnotesize
\setlength{\tabcolsep}{1.5pt}
\begin{tabularx}{\linewidth}{ccc|>{\centering\arraybackslash}X|*{5}{>{\centering\arraybackslash}X}|*{4}{>{\centering\arraybackslash}X}}
\toprule
\multicolumn{13}{c}{\textbf{LingBot-Video (30B-A3B)}} \\
\midrule
\multicolumn{3}{c|}{Components} & Overall & \multicolumn{5}{c|}{Task Scores$\uparrow$} & \multicolumn{4}{c}{Embodiment Scores$\uparrow$} \\
\cmidrule(lr){1-3}\cmidrule(lr){4-4}\cmidrule(lr){5-9}\cmidrule(lr){10-13}
(1) & (2) & (3) & Avg$\uparrow$ & Manip. & Spatial & Multi. & Long. & Reason. & Single. & Dual. & Quad. & Hum. \\
\midrule
-- & -- & -- & 0.805 & 0.898 & 0.860 & 0.898 & 0.908 & 0.912 & 0.678 & 0.671 & 0.681 & 0.743 \\
\midrule
$\checkmark$ & -- & -- & 0.827 & 0.925 & 0.860 & 0.880 & 0.940 & 0.915 & 0.709 & \textbf{0.706} & 0.746 & 0.766 \\
$\checkmark$ & $\checkmark$ & -- & 0.828 & 0.928 & 0.840 & \textbf{0.902} & 0.948 & 0.932 & \textbf{0.711} & 0.678 & 0.746 & 0.768 \\
\midrule
$\checkmark$ & $\checkmark$ & $\checkmark$ & \textbf{0.835} & \textbf{0.940} & \textbf{0.860} & 0.880 & \textbf{0.959} & \textbf{0.945} & 0.692 & 0.702 & \textbf{0.748} & \textbf{0.786} \\
\bottomrule
\end{tabularx}
\end{table}

\pagebreak
\begingroup
\raggedbottom
\section{Qualitative Comparisons}
\label{app:qualitative-comparisons}
\begin{figure}[H]
\centering
\includegraphics[page=1,width=\linewidth]{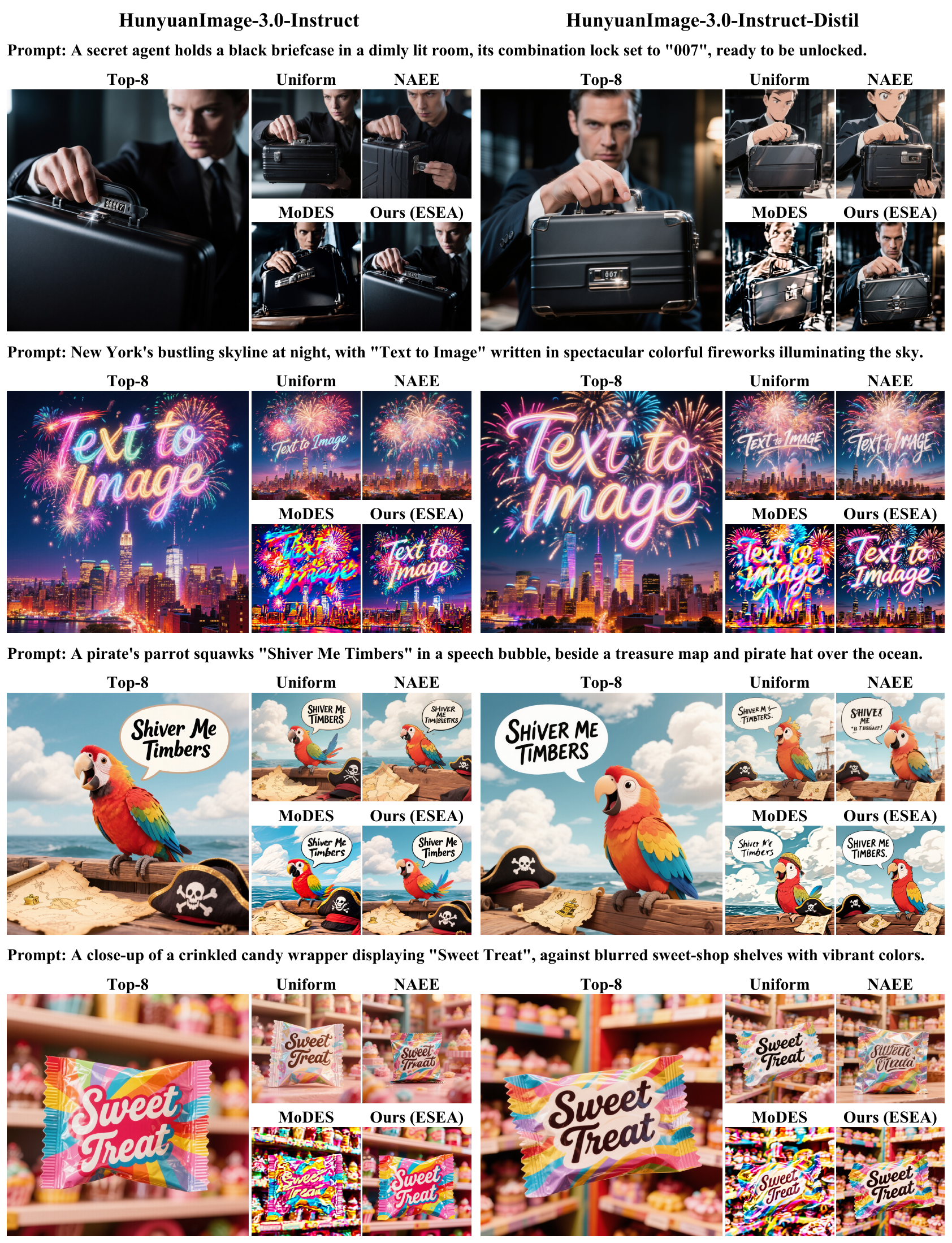}
\caption{Qualitative comparisons on HunyuanImage-3.0-Instruct (left) and its distilled variant (right).}
\label{fig:qualitative-ocr}
\end{figure}

\begin{figure}[H]
\centering
\includegraphics[page=1,width=0.92\linewidth]{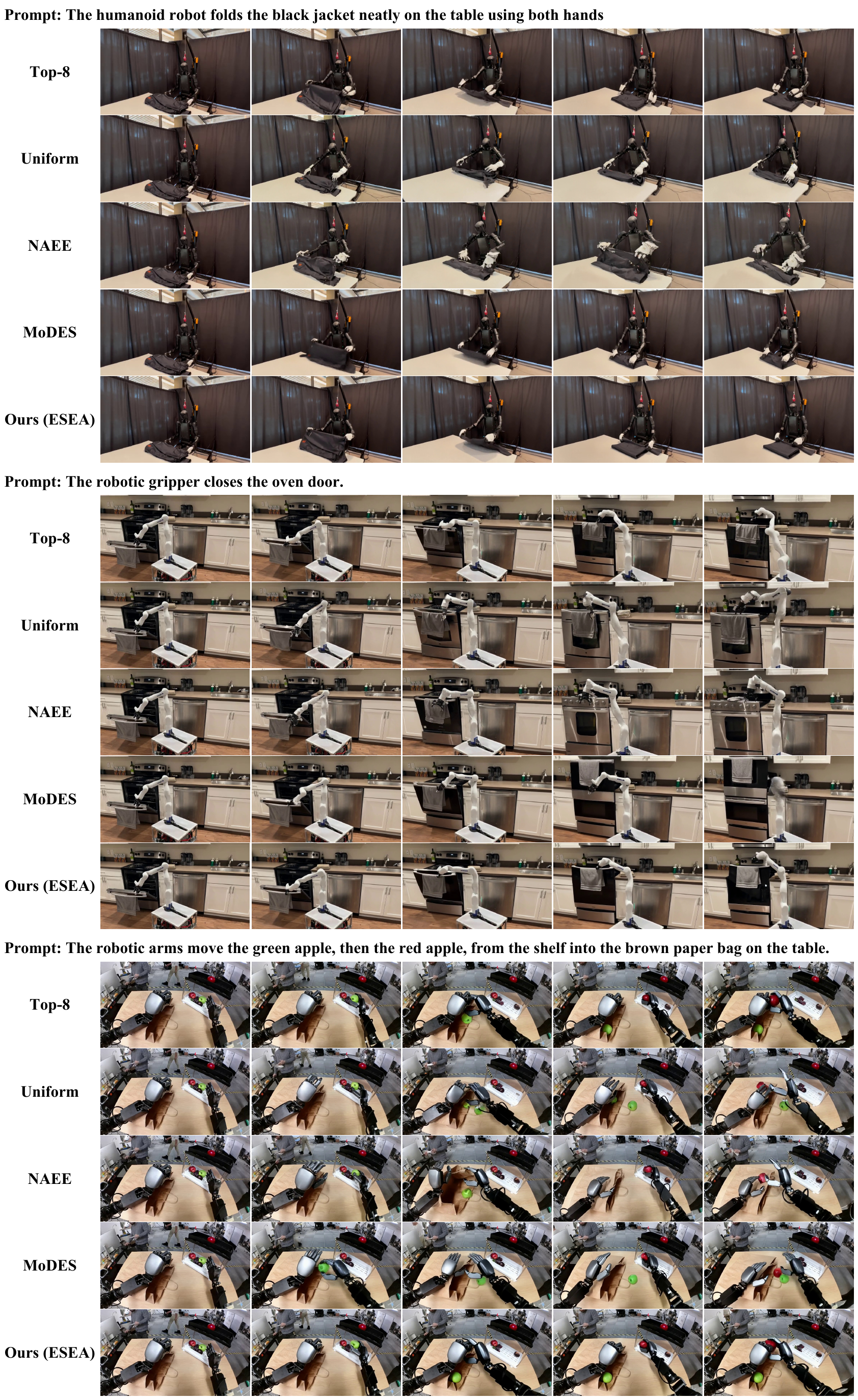}
\caption{Qualitative comparisons. The frames are uniformly sampled from the generated video.}
\label{fig:qualitative-lingbot}
\end{figure}

\par\endgroup

\end{document}